\documentclass[a4paper,11pt]{article}

\usepackage{fourier}
\usepackage{color}
 \usepackage{graphicx}
\usepackage{url}
\usepackage[affil-it]{authblk}
\usepackage{amsmath}
\usepackage{wrapfig}

\usepackage[T1]{fontenc}
\usepackage{times}

\begin{document}
\title{A Comparative Study of Image-to-Image Translation Using GANs for Synthetic Child Race Data}

\author[1]{Wang Yao} 
\author[1]{Muhammad Ali Farooq}
\author[2]{Joseph Lemley}
\author[1]{Peter Corcoran}

\affil[1]{School of Engineering, University of Galway, Ireland.}
\affil[2]{Xperi Corporation, Galway.}

\date{}
\maketitle
\thispagestyle{empty}

\vspace{-30pt}
\begin{abstract}
The lack of ethnic diversity in data has been a limiting factor of face recognition techniques in the literature. This is particularly the case for children where data samples are scarce and presents a challenge when seeking to adapt machine vision algorithms that are trained on adult data to work on children. This work proposes the utilization of image-to-image transformation to synthesize data of different races and thus adjust the ethnicity of children's face data. We consider ethnicity as a style and compare three different Image-to-Image neural network based methods, specifically pix2pix, CycleGAN, and CUT networks to implement Caucasian child data and Asian child data conversion. Experimental validation results on synthetic data demonstrate the feasibility of using image-to-image transformation methods to generate various synthetic child data samples with broader ethnic diversity. 
\end{abstract}

\section{Introduction}
In recent years face authentication has witnessed significant advancements and has become widely deployed in various applications, such as authentication systems, immigration management, and financial security. Studies~\cite{abdurrahim2018review, cavazos2020accuracy} have shown that these face authentication systems exhibit biases, especially when it comes to recognizing faces from certain racial or ethnic groups. This may lead to discrimination and injustice against specific groups, for instance by wrongly identifying them as suspects, restricting their access rights, or other such issues. Thus, race imbalance is a pressing issue that demands attention in face authentication applications.

However, collecting large amounts of effective race/ethnicity data in the real world is laborious and challenging because the process of data acquisition is expensive and time-consuming, especially when it comes to human subjects. Considering  the General Data Protection Regulations (GDPR) in European Union (EU) region~\cite{EuropeanParliament2016a}, when collecting any video, image, or audio data from human subjects, the scope of usage of such data and any subsequent processing of the data must be clearly defined and explained to the subject, which normally requires explicit consent. In addition, it is important that data must be stored securely and that it supports a series of rights, for example, the right of the subject to retract the stored data at any time. This becomes more complex in the case of engaging with child subjects, as the consent of the legal guardian is required, and it is preferable to inform the subject in plain language about the scope of collecting and further using this data.

Our work with the DAVID smart-toy platform~\cite{davidproject} has motivated us to explore the generation of synthetic facial data, which is not subject to data protection regulations. In this work, we consider ethnicity as a style and employ style-to-style transformation to synthesize data from different races. This research adopts the potential of Image-to-Image (I2I) translation approaches to generate synthetic child race data, which will benefit the diversity of training data, reducing the ethnic bias of facial recognition systems, and improving the robustness of machine vision algorithms trained on this type of synthetic data. In this work, we aim to generate synthetic Caucasian child data and Asian child data by training three I2I translation methods including pix2pix~\cite{isola2017image}, CycleGAN~\cite{zhu2017unpaired}, and CUT~\cite{park2020contrastive}. Then we qualitatively and quantitatively validated the synthetic child racial samples using machine vision algorithms.

\section{Related Works}

Image-to-Image translation methods refer to converting input images from a source domain to a target domain while preserving the content representations of the input image. I2I algorithms can solve many problems in computer vision tasks, such as image registration, image segmentation, and image restoration. It can be categorized into supervised I2I and unsupervised I2I based on whether the source domain images and target domain images are aligned image pairs. Isola et al.~\cite{isola2017image} proposed pix2pix to solve various supervised I2I tasks by adopting a conditional GAN framework, which is also a baseline for the image translation framework. However, training supervised translation in real-world scenes is impractical because it is difficult to create a paired dataset. CycleGAN~\cite{zhu2017unpaired} and its variants such as TraVeLGAN~\cite{amodio2019travelgan} employ cycle-consistency loss, which was proven effective in solving this problem. Study~\cite{pang2021image} reveals that most methods using cycle consistency constraints tend to directly synthesize a new domain with a global target style translation and rarely consider local objects or fine-grained instances during translation. CUT~\cite{park2020contrastive} employs contrastive learning in a patch-based way instead of learning the entire images. LPTN~\cite{liang2021high} uses a Laplacian pyramid to decompose the input and achieve I2I translation.

Even though these I2I methods perform well in many tasks, one major challenge of these approaches is to achieve robust results for race-to-race child facial transformation applications. It is difficult to find a large-scale real child dataset with multiple race classes. Existing large-scale face datasets such as VGGFace2~\cite{cao2018vggface2}, MS-Celeb-1M~\cite{guo2016ms}, and FFHQ~\cite{karras2019style} are generally focused on adult data. Although children's faces are present in some age datasets, the amount of data is small and the resolution varies widely~\cite{moschoglou2017agedb, zhifei2017cvpr}. Moreover, few studies have been conducted to generate synthetic data for different races~\cite{yucer2020exploring, ba2021overcoming}. One study proposed by Yucer et al.~\cite{yucer2020exploring} is focused on generating synthetic race data through CycleGAN and using this data to improve face recognition accuracy. Another recent study~\cite{ba2021overcoming} generates synthetic skin tones while retaining their pulsatile signals for exploring physiological signals. Both studies are focused on generating adult data. In this work, we explore generating synthetic child race data by using I2I translation methods.

\section{Methodology}
\begin{wrapfigure}{r}{0.4\textwidth}
  \vspace{-40pt}
  \begin{center}
    \includegraphics[width=0.35\textwidth, height=0.35\textwidth]{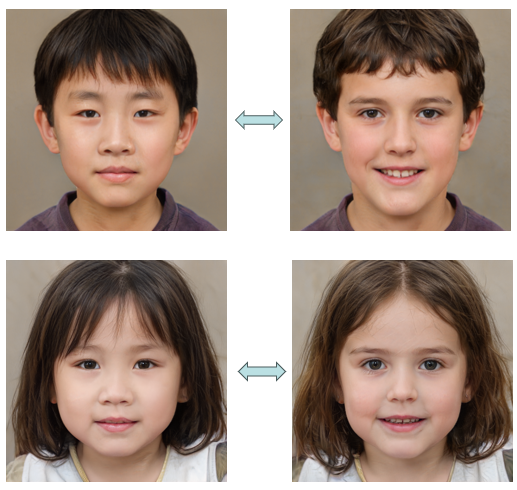}      
   \end{center}
\vspace{-15pt}
\caption{Example of two image pairs.}
\label{fig1}
\vspace{-30pt}
\end{wrapfigure}
In this section, datasets, I2I translation methods, and evaluation metrics that were utilized in this study are detailed.
\subsection{Datasets}

In this work, we collect a synthetic child race dataset by finetuning a pre-trained StyleGAN2~\cite{karras2020analyzing} network. This dataset comprises of data from 2400 Asian children  ($1200$ girls $+ 1200$ boys) and 2400 Caucasian children $(1200$ girls $+ 1200$ boys), which are generated by using the latent space editing technique~\cite{wu2021stylespace}. We divided the dataset into groups of boys and girls and paired Asian children with Caucasian children one by one. The paired examples are shown in Figure~\ref{fig1}. 

To the best of our knowledge, there is no existing large-scale dataset focused on child race facial data. As mentioned in related work, recording real-world large-scale child race data is a laborious task that will also require ethical approvals and must be performed according to GDPR regulations therefore in this work we have mainly focused on using synthetic child data for preserving personal identification data (PID). Moreover, considering the comparison with the supervised I2I method, we want to collect race data in image pairs. For this purpose, we have focused on using latent space editing in StyleGAN2~\cite{karras2020analyzing} to get the image pairs. Our goal is
to generate child facial data with diversified ethnicity. To achieve this we have first fine-tuned StyleGAN2 \cite {karras2020analyzing} to produce synthetic child images in previous work~\cite{farooq2023childgan}. This paper serves as an initial study to compare the
feasibility of different I2I methods for synthesizing child racial face data.

\subsection{I2I translation methods}

In this work, we have incorporated three typical I2I translation methods which are discussed below. 

\paragraph{Pix2pix:} The pix2pix model~\cite{isola2017image} is a conditional GAN, which means that the output image is generated conditionally on the input image. The discriminator gets the source image and the target image and determines whether the target image is a reasonable transformation of the source image. It is a supervised I2I model and requires many aligned image pairs for training. 

\paragraph{CycleGAN:} CycleGAN~\cite{zhu2017unpaired} consists of two generators and two discriminators and designed cycle-consistent adversarial networks for unpaired I2I translation. A cycle-consistency loss is designed to measure the difference between the synthetic image produced by the second generator and the source image. CycleGAN uses cycle-consistent constraints to train unsupervised image translation models through the GAN architecture, which enables the conversion between two unpaired image sets.

\paragraph{CUT:} CUT~\cite{park2020contrastive} is unsupervised one-side translation beyond cycle-consistency constraint. It utilizes a contrastive learning framework to maximize the mutual information between two patches. A multi-layer, patch-based approach is used to encourage the generated images to be similar to the source images. This model avoids the use of cycle-consistency loss, and only one set of GANs is needed for image transformation.

\subsection{Evaluation metrics}
Three well-knowon quantitative evaluation metrics have been selected for validating the synthetic child race facial data. These are listed below with a brief discussion on what image quality each of these measures. 

\paragraph{FID:} Fréchet inception distance (FID)~\cite{heusel2017gans} calculates the distance of the distribution between the synthetic images and the source images. The lower FID score means the model has a better performance.
\paragraph{PSNR:} Peak signal-to-noise ratio (PSNR) measures the intensity differences between the reference image and the test image. The higher PSNR score indicates a higher quality of the test image.
\paragraph{SSIM:} Structural similarity index (SSIM)~\cite{wang2004image} computes the perceptual distance between the reference image and the test image according to luminance, contrast, and structure. The higher SSIM score implies that the greater the similarity between the reference image and the test image.

\begin{table}[!h]
\begin{center}
\begin{tabular}{|c|c|c|c|}
\hline
\textbf{S.no} & \textbf{Parameter}        & \textbf{Pix2Pix/CycleGAN Value} & \textbf{CUT Value} \\ \hline
1             & Preprocess                & Scale width                     & Scale width        \\ \hline
2             & Load size                 & 256                             & 256                \\ \hline
3             & Batch size                & 8                               & 8                  \\ \hline
4             & Learning Rate             & 0.0002                          & 0.0002             \\ \hline
5             & Learning Rate Policy      & linear                          & linear             \\ \hline
6             & Learning Rate Decay Iters & 50                              & 50                 \\ \hline
7             & Dropout                   & False                           & False              \\ \hline
8             & Discriminator             & PatchGAN                        & PatchGAN           \\ \hline
9             & Generator                 & resnet\_9blocks                 & resnet\_9blocks    \\ \hline
10            & Mirror augment            & False                           & True               \\ \hline
11            & Training epochs           & 200                             & 400                \\ \hline
\end{tabular}
\caption{Hyper-parameter Selection}
\label{tab:tab1}
\vspace{-20pt}
\end{center}
\end{table}

\begin{figure}[htb]
\centering
\includegraphics[width=0.48\linewidth]{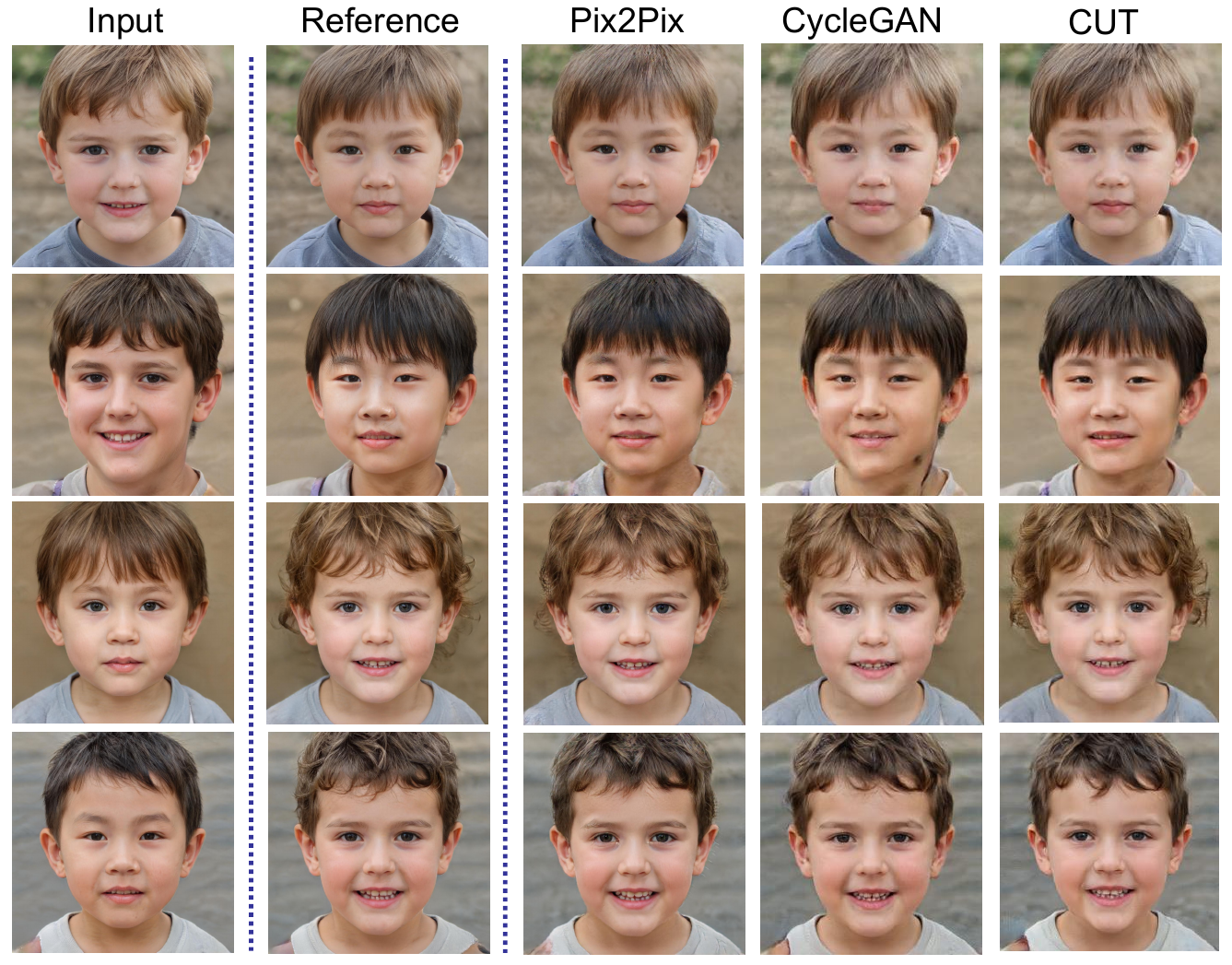}
\includegraphics[width=0.48\linewidth]{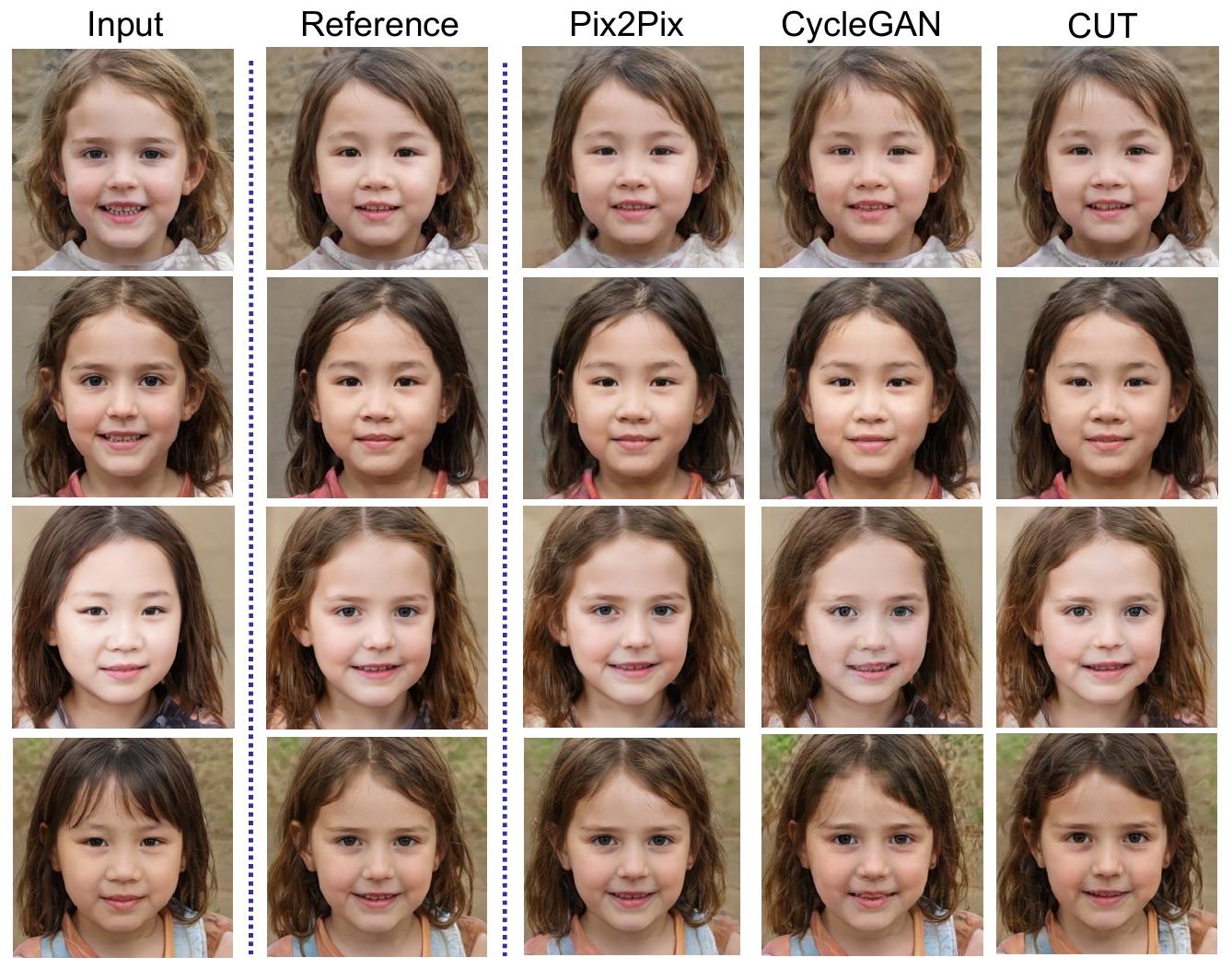}
\caption{Qualitative results of synthetic boys and girls race data.}
\label{fig2}
\end{figure}
\section{Experiment}

\paragraph{Training Setting:} Four mappings \{Caucasian boy → Asian boy, Asian boy → Caucasian boy, Caucasian girl → Asian girl, Asian girl → Caucasian girl\} are adopted in our work. As pix2pix and CUT uses one directional transformation, four models are trained separately during the experiments. CycleGAN has bidirectional transformations, and  thus only two models are trained during training. All the child faces are aligned and resized to $256 \times 256$. Table~\ref{tab:tab1} shortlisted the optimal set of network hyperparameters during the training process. The complete experiment was performed on a server-grade machine equipped with 2 NVIDIA GeForce GTX TITAN.
\subsection{Qualitative Evaluation}
\begin{wrapfigure}{r}{0.5\textwidth}
  \vspace{-55pt}
  \begin{center}
    \includegraphics[width=0.40\textwidth]{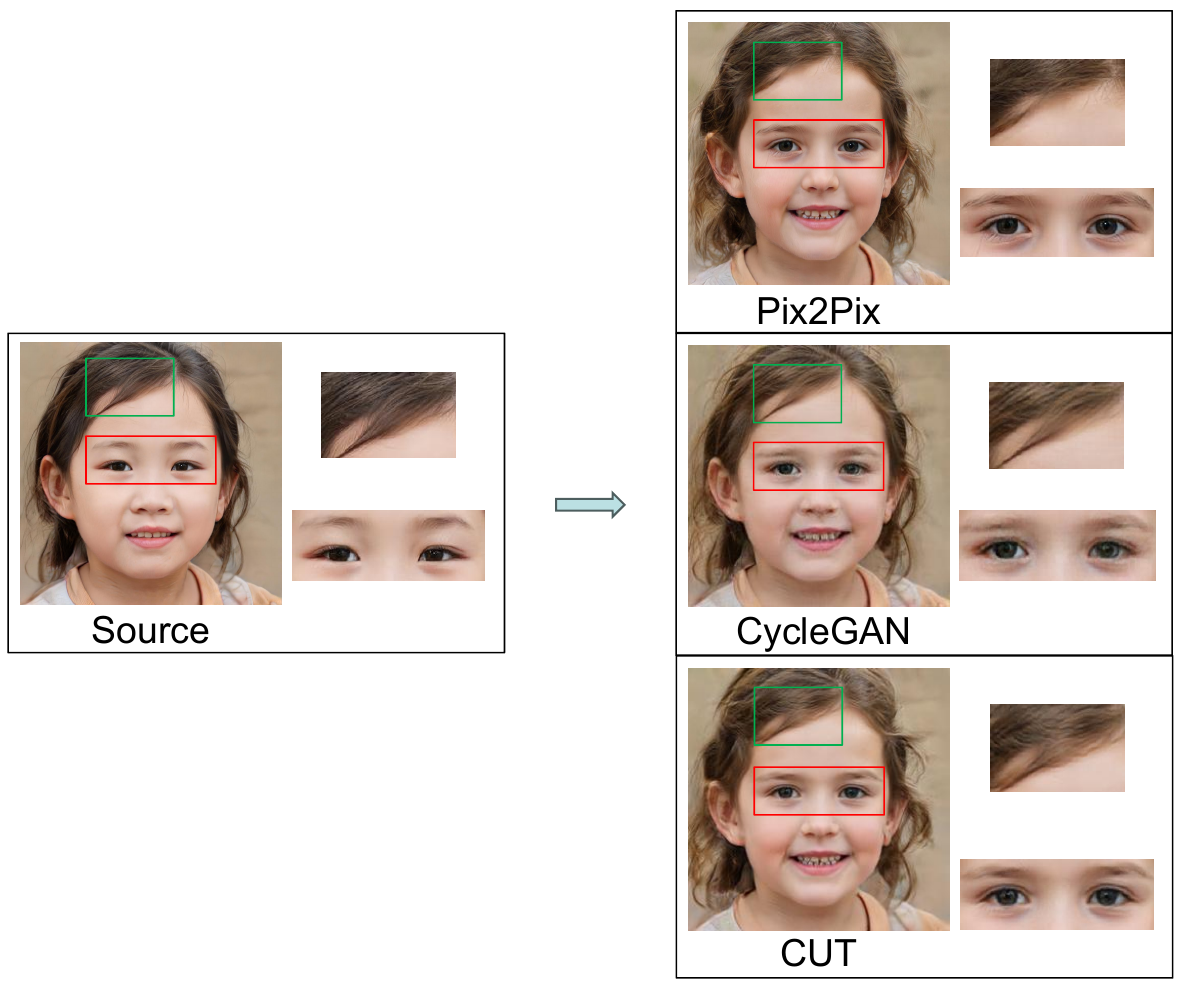} 
\end{center}
\vspace{-20pt}
\caption{Example of converting an Asian girl to a Caucasian girl}
\label{fig3}
  \vspace{-20pt}
\end{wrapfigure}
Figure~\ref{fig2} shows the qualitative evaluation results from three I2I translation methods. The first two rows show the conversion from Caucasian to Asian. The last two rows show the conversion from Asian to Caucasian. The results from pix2pix show the best visual results, which are most similar to the reference image. The next most effective result is the image synthesized by the CUT.
Figure~\ref{fig3} shows the details of different facial features of generated girl’s Caucasian faces. The shape of the eyes has changed significantly in Figure~\ref{fig3}, especially for the girl image generated by pix2pix. Secondly, we can observe the hair color transformation since it becomes light brown and the shape of the hair is more curly when compared to the original (reference) image.

We will conduct a MOS study to provide further evidence on the quality of generated images but are currently waiting for ethical approvals. In the meantime, several people in our group have assisted with a small informal study where they are asked to distinguish between the original data and the generated data. From the results of this informal study, indications are that synthetic data are difficult to distinguish from original data samples. 

\subsection{Quantitative Evaluation}

In order to evaluate the performance of these tuned GAN models, we have generated 100 Caucasian boys, 100 Asian boys, 100 Caucasian girls, and 100 Asian girls through Pix2Pix, CycleGAN, and CUT using unseen test data. We have calculated the average FID, PSNR, and SSIM scores through the reference images and generated images. The results are shown in Table~\ref{tab:tab2}. The synthetic images generated from pix2pix have the highest scores on PSNR and SSIM, which indicate that the synthetic images generated from pix2pix comprises robust quality facial features from human perception. CUT has the best FID value through three models, which means the distribution of generated images from CUT is close to the distribution of the source images.

\subsection{Synthetic Facial Race Analysis}

To analyze the ethnicity attribute of synthetic child data, a pre-trained Deepface~\cite{serengil2021lightface} model is used to classify the race of generated child data. This model is trained on a large-scale real balanced ethnicity dataset, which has six race labels including Asian, white, middle eastern, Indian, Latino Hispanic, and black. Since this work focuses on two types of race data i.e. Asian and Caucasian/white; thus we only focus on the classification accuracy of Asian and Caucasian/white. Table~\ref{tab:tab3} shows the results of the racial classification. From Table~\ref{tab:tab3}, it can be observed that synthetic child race data were classified with robust accuracy levels of 99\% for Asian boys using pix2pix and CUT whereas we achieved 97\% for Caucasian girls using pix2pix.

\begin{table}[]
\centering
\begin{tabular}{|cccc|}
\multicolumn{4}{c}{Asian Boy $\rightarrow$ Caucasian Boy}                                                                                       \\ \hline
\multicolumn{1}{|c|}{\textbf{Evaluation Metrics}} & \multicolumn{1}{c|}{\textbf{Pix2Pix}~\cite{isola2017image} } & \multicolumn{1}{c|}{\textbf{CycleGAN}~\cite{zhu2017unpaired}} & \textbf{CUT}~\cite{park2020contrastive} \\ \hline
\multicolumn{1}{|c|}{\textbf{FID $(\downarrow)$}}              & \multicolumn{1}{c|}{49.59}            & \multicolumn{1}{c|}{31.46}             & 26.36        \\ \hline
\multicolumn{1}{|c|}{\textbf{PSNR $(\uparrow)$}}               & \multicolumn{1}{c|}{24.41}            & \multicolumn{1}{c|}{22.09}             & 22.10        \\ \hline
\multicolumn{1}{|c|}{\textbf{SSIM $(\uparrow)$}}               & \multicolumn{1}{c|}{0.75}             & \multicolumn{1}{c|}{0.65}              & 0.65         \\ \hline
\multicolumn{4}{c}{Caucasian Boy $\rightarrow$ Asian Boy}                                                                                       \\ \hline
\multicolumn{1}{|c|}{\textbf{Evaluation Metrics}} & \multicolumn{1}{c|}{\textbf{Pix2Pix}~\cite{isola2017image} } & \multicolumn{1}{c|}{\textbf{CycleGAN}~\cite{zhu2017unpaired}} & \textbf{CUT}~\cite{park2020contrastive} \\ \hline
\multicolumn{1}{|c|}{\textbf{FID $(\downarrow)$}}              & \multicolumn{1}{c|}{29.62}                & \multicolumn{1}{c|}{29.54}             &  25.06        \\ \hline
\multicolumn{1}{|c|}{\textbf{PSNR $(\uparrow)$}}               & \multicolumn{1}{c|}{25.38}                & \multicolumn{1}{c|}{22.64}             &  22.74            \\ \hline
\multicolumn{1}{|c|}{\textbf{SSIM $(\uparrow)$}}               & \multicolumn{1}{c|}{0.78}                 & \multicolumn{1}{c|}{0.68}              &  0.67            \\ \hline
\multicolumn{4}{c}{Asian Girl $\rightarrow$ Caucasian Girl}                                                                                     \\ \hline
\multicolumn{1}{|c|}{\textbf{Evaluation Metrics}} & \multicolumn{1}{c|}{\textbf{Pix2Pix}~\cite{isola2017image} } & \multicolumn{1}{c|}{\textbf{CycleGAN}~\cite{zhu2017unpaired}} & \textbf{CUT}~\cite{park2020contrastive} \\ \hline
\multicolumn{1}{|c|}{\textbf{FID $(\downarrow)$}}                & \multicolumn{1}{c|}{26.31}                 & \multicolumn{1}{c|}{42.27}              &  25.09      \\ \hline
\multicolumn{1}{|c|}{\textbf{PSNR $(\uparrow)$}}               & \multicolumn{1}{c|}{24.02}                 & \multicolumn{1}{c|}{21.42}                &  21.58            \\ \hline
\multicolumn{1}{|c|}{\textbf{SSIM $(\uparrow)$}}               & \multicolumn{1}{c|}{0.70}                 & \multicolumn{1}{c|}{0.60}                  &  0.62           \\ \hline
\multicolumn{4}{c}{Caucasian Girl $\rightarrow$ Asian Girl}                                                                                     \\ \hline
\multicolumn{1}{|c|}{\textbf{Evaluation Metrics}} & \multicolumn{1}{c|}{\textbf{Pix2Pix}~\cite{isola2017image} } & \multicolumn{1}{c|}{\textbf{CycleGAN}~\cite{zhu2017unpaired}} & \textbf{CUT}~\cite{park2020contrastive} \\ \hline
\multicolumn{1}{|c|}{\textbf{FID $(\downarrow)$}}                & \multicolumn{1}{c|}{44.47}                 & \multicolumn{1}{c|}{43.24}              &  34.87       \\ \hline
\multicolumn{1}{|c|}{\textbf{PSNR $(\uparrow)$}}               & \multicolumn{1}{c|}{24.29}                 & \multicolumn{1}{c|}{21.15}                &  21.73            \\ \hline
\multicolumn{1}{|c|}{\textbf{SSIM $(\uparrow)$}}               & \multicolumn{1}{c|}{0.70}                 & \multicolumn{1}{c|}{0.62}                  &   0.63           \\ \hline
\end{tabular}
\caption{The average FID, PSNR, and SSIM scores of different I2I methods.}
\label{tab:tab2}
\end{table}

\begin{table}[]
\centering
\begin{tabular}{|c|c|c|c|c|}
\hline 
\textbf{Model}    & \textbf{Test Data}      & \ \ \textbf{Asian}\ \  & \textbf{Caucasian} & \textbf{Other Race} \\ \hline
pix2pix  & Caucasian Boy  & 4\%    & 76\%       & 20\%        \\ 
CycleGAN & Caucasian Boy  & 5\%    & 76\%       & 19\%        \\ 
CUT      & Caucasian Boy  & 8\%    & 72\%       & 20\%        \\ \hline
pix2pix  & Asian Boy      & 99\%   & 1\%        & 0          \\ 
CycleGAN & Asian Boy      & 96\%   & 2\%        & 2\%         \\ 
CUT      & Asian Boy      & 99\%   & 1\%        & 0          \\ \hline
pix2pix  & Caucasian Girl & 0     & 97\%       & 3\%         \\ 
CycleGAN & Caucasian Girl & 4\%    & 85\%       & 11\%        \\ 
CUT      & Caucasian Girl & 0     & 94\%       & 6\%         \\ \hline
pix2pix  & Asian Girl     & 89\%   & 10\%       & 1\%         \\ 
CycleGAN & Asian Girl     & 96\%   & 3\%        & 1\%         \\ 
CUT      & Asian Girl     & 89\%   & 9\%        & 2\%         \\ \hline 
\end{tabular}
\caption{The average accuracy of race classification.}
\label{tab:tab3}
\end{table}
\section{Discussion}

In this work, we have focused on using synthetic child data for generating different race variations. The main reason for us to explore using synthetic data over real child data is due to the DAVID embedded smart toy platform~\cite{davidproject} where child data is required to fine-tune and compress computer vision models originally optimized on adult subjects. In this project, we have tried to gather child data by using a 3D scanner. When we engaged with the data protection office, we realized that the complexity of managing and providing access to original child data made this approach impractical. 

This is reflected in several aspects as below.

\begin{itemize}
\item Personally Identifiable Data (PID) Protection: Synthetic data can be used to avoid collecting sensitive personally identifiable Data (PID) from a vulnerable population of data subject, such as children. State-of-the-art data synthesis techniques enable data from adult subjects to be adapted to generate corresponding child data samples and in future work we can validate these data against original dataset when, and if, these become publicly available.
\item Cost Effective: Utilizing synthetic data methods is far more cost-effective when compared to collecting and labelling large volumes of real child data. Collecting real data can involve significant expenses, such as data collection infrastructure, data storage, and data labelling costs. Thus, synthetic data generation and further utilizing it for experimental analysis avoids these costs, making it an attractive option for training advanced machine learning models.
\item Controllable Data: Our work provides a range of tools, based on state-of-the-art data transformation models, to add expression, lighting, age, gender and pose variations to the original data samples. This work adds the additional capability to provide ethnic variations, an important tools to help diversify machine vision algorithms based on neural-AI models.   
\item GDPR Compliance: In situations where it is necessary to use and further share the data for research analysis while ensuring the anonymity of individuals, synthetic data can be freely employed. By replacing real data with synthetic data, the privacy of individuals can be preserved, thus addressing ethical concerns and complying with general data protection regulations (GDPR).
\end{itemize}

Note that a full validation of our proposed use of models based on generated data will require testing on original child subjects. This work is part of the DAVID roadmap and one of our industry partners has collected suitable original data from c.500 child subjects. Unfortunately, due to GDPR, such data can only be shared within the DAVID research consortium.   


\section{Conclusion}
We have presented a comparative study on image-to-image translation methods including pix2pix, CycleGAN, and CUT to generate child race data which aims to solve the diversity issues in child data. As an initial work, we have trained ten models to explore the translation between Caucasian child faces and Asian child faces in this research and the experimental results show that it is feasible to synthesize child race faces through image-to-image translation. Three evaluation metrics have been adopted in our experiment. The results show that CUT has the highest FID value of all models, while pix2pix has the highest PSNR and SSIM scores. 

A pretrained ethnic classification model is introduced to evaluate the synthetic race data, which shows that synthetic child ethnicity can be classified accurately. Since the dataset used in our work was manipulated by the latent code of a finetuned StyleGAN2 model, each pair of images (Asian $\leftrightarrow$ Caucasian) has some underlying similarity. This may introduce some limitations but this work should be regarded as early-stage proof-of-concept rather than a comprehensive study. Future work will extend to handle more diversified race generation, and combine this work with state-of-the-art text-to-image frameworks.

\section*{Acknowledgments}
\vspace{-5pt}
This research is supported by (i) Irish Research Council Enterprise Partnership Ph.D. Scheme (Project ID: EPSPG/2020/40), (ii) Xperi Corporation, Ireland, and (iii) the Data-Center Audio/Visual Intelligence on-Device (DAVID) Project (2020–2023) funded by the  Disruptive Technologies Innovation Fund (DTIF).
\vspace{-5pt}

\appendix

\bibliographystyle{apalike}

\bibliography{imvip}

\end{document}